\documentclass{ieeeaccess}
\usepackage{cite}
\usepackage{amsmath,amssymb,amsfonts}
\usepackage{algorithmic}
\usepackage{graphicx}
\usepackage{textcomp}
\usepackage{caption}
\usepackage{subcaption}
\usepackage{multirow}
\usepackage{booktabs}
\usepackage{url}
\def\BibTeX{{\rm B\kern-.05em{\sc i\kern-.025em b}\kern-.08em
    T\kern-.1667em\lower.7ex\hbox{E}\kern-.125emX}}
    
\begin{document}

\history{Date of publication xxxx 00, 0000, date of current version xxxx 00, 0000.}
\doi{10.1109/ACCESS.2024.DOI}

\title{YOLOv8-AM: YOLOv8 Based on Effective Attention Mechanisms for Pediatric Wrist Fracture Detection}

\author{
\uppercase{Chun-Tse Chien}\authorrefmark{1}, 
\uppercase{Rui-Yang Ju}\authorrefmark{2}, \IEEEmembership{Student Member, IEEE},
\uppercase{Kuang-Yi Chou}\authorrefmark{3},
\uppercase{Enkaer Xieerke}\authorrefmark{4},
\uppercase{Jen-Shiun Chiang}\authorrefmark{1}, \IEEEmembership{Member, IEEE}}

\address[1]{Department of Electrical and Computer Engineering, Tamkang University, New Taipei City, 251301, Taiwan}
\address[2]{Graduate Institute of Networking and Multimedia, National Taiwan University, Taipei City 106335, Taiwan}
\address[3]{School of Nursing, National Taipei University of Nursing and Health Sciences, Taipei City, 112303, Taiwan}
\address[4]{College of Energy and Mechanical Engineering, Shanghai University of Electric Power, Shanghai, 201306, China}

\tfootnote{This paper is an expanded paper from International Conference on Neural Information Processing (ICONIP) held on December 2-6, 2024 in Auckland, New Zealand. 
This work is supported by National Science and Technology Council of Taiwan, under Grant Number: NSTC 112-2221-E-032-037-MY2.}

\markboth
{Chien \headeretal: YOLOv8-AM}
{Chien \headeretal: YOLOv8-AM}

\corresp{Corresponding author: Jen-Shiun Chiang 
(e-mail: jsken.chiang@gmail.com).}

\begin{abstract}
Wrist trauma and even fractures occur frequently in daily life, particularly among children who account for a significant proportion of fracture cases.
Before performing surgery, surgeons often request patients to undergo X-ray imaging first and prepare for it based on the analysis of the radiologist. 
With the development of neural networks, You Only Look Once (YOLO) series models have been widely used in fracture detection as computer-assisted diagnosis (CAD). 
In 2023, Ultralytics presented the latest version of the YOLO models, which has been employed for detecting fractures across various parts of the body. 
Attention mechanism is one of the hottest methods to improve the model performance. 
This research work proposes YOLOv8-AM, which incorporates the attention mechanism into the original YOLOv8 architecture. 
Specifically, we respectively employ four attention modules, Convolutional Block Attention Module (CBAM), Global Attention Mechanism (GAM), Efficient Channel Attention (ECA), and Shuffle Attention (SA), to design the improved models and train them on GRAZPEDWRI-DX dataset. 
Experimental results demonstrate that the mean Average Precision at IoU 50 (mAP 50) of the YOLOv8-AM model based on ResBlock + CBAM (ResCBAM) increased from 63.6\% to 65.8\%, which achieves the state-of-the-art (SOTA) performance. 
Conversely, YOLOv8-AM model incorporating GAM obtains the mAP 50 value of 64.2\%, which is not a satisfactory enhancement. 
Therefore, we combine ResBlock and GAM, introducing ResGAM to design another new YOLOv8-AM model, whose mAP 50 value is increased to 65.0\%. 
The implementation code for this study is available on GitHub at \url{https://github.com/RuiyangJu/Fracture_Detection_Improved_YOLOv8}.
\end{abstract}


\titlepgskip=-15pt

\maketitle
\section{Introduction}
Wrist fractures are one of the most common fractures, particularly among the elderly and children \cite{hedstrom2010epidemiology,randsborg2013fractures}. 
Fractures typically occur in the distal 2 cm of the radius near the joint. 
Failure to provide timely treatment may result in deformities of the wrist joint, restricted joint motion, and joint pain for the patients \cite{bamford2010qualitative}. 
In children, a misdiagnosis would lead to a lifelong inconvenience \cite{kraus2010treatment}.

\begin{figure*}[ht]
  \centering
  \includegraphics[width=\linewidth]{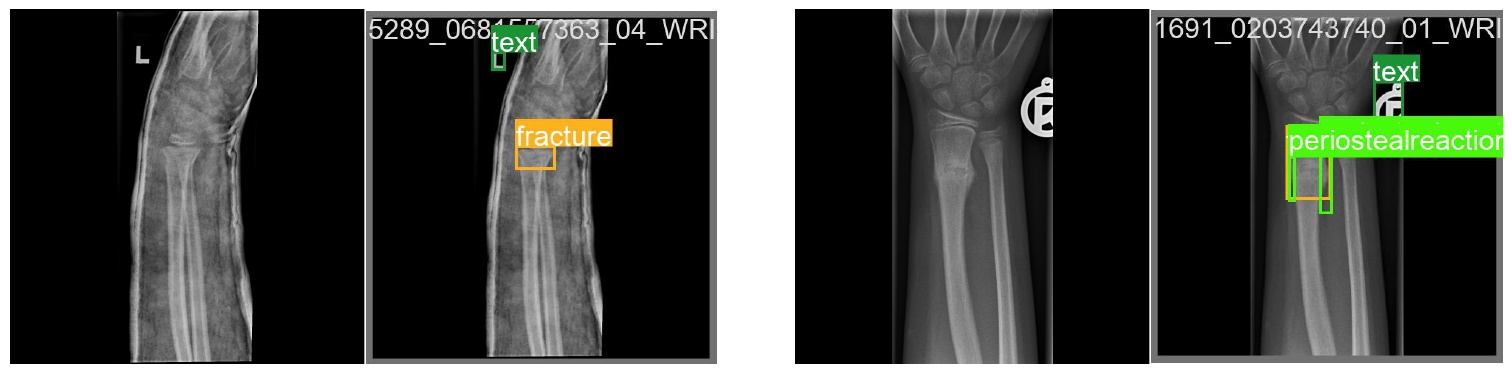}
  \caption{Example diagrams illustrating the pediatric wrist fracture prediction task conducted in this work, with the input image displayed on the left of each group and the prediction result shown on the right of each group.}
  \label{fig_task}
\end{figure*}

\begin{figure*}[ht]
  \centering
  \includegraphics[width=\linewidth]{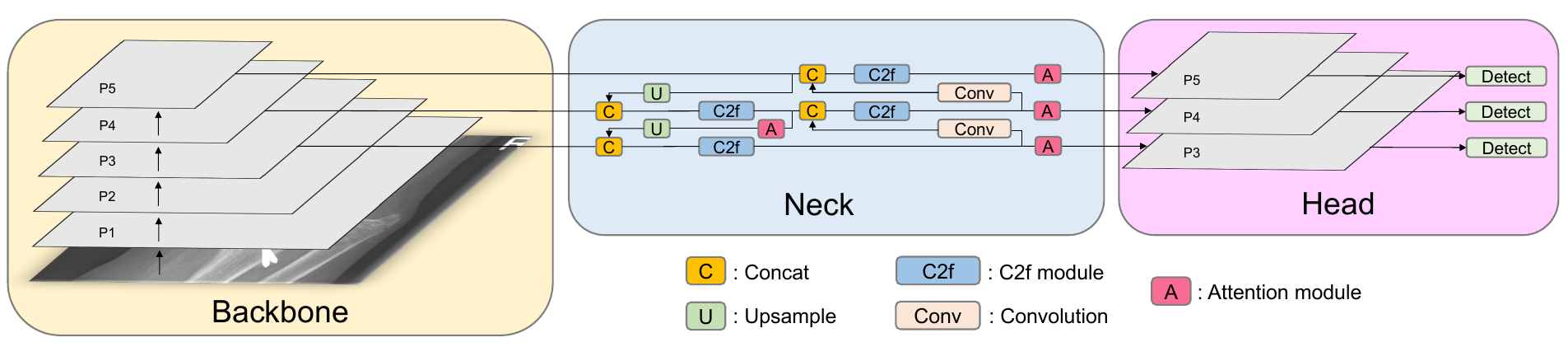}
  \caption{The architecture of the YOLOv8-AM model for pediatric wrist fracture detection, where the attention module is the part newly employed to the original YOLOv8 model architecture.}
  \label{fig_overall}
\end{figure*}

In cases of pediatric wrist fractures, surgeons inquire about the circumstances leading to the fracture and conduct a preoperative examination. 
Presently, fracture examinations mainly utilize three types of medical imaging equipment: 
Magnetic Resonance Imaging (MRI), Computed Tomography (CT), and X-ray. 
Among these, X-ray is the preferred choice for most patients due to its cost-effectiveness \cite{wolbarst1999looking}. 
In hospitals providing advanced medical care, radiologists are required to follow the Health Level 7 (HL7) and Digital Imaging and Communications in Medicine (DICOM) international standards for the archival and transfer of X-ray images \cite{boochever2004his}. 
Nevertheless, the scarcity of radiologists in underdeveloped regions poses a significant challenge to the prompt delivery of patient care \cite{burki2018shortfall,rimmer2017radiologist,rosman2015imaging}. 
The studies \cite{erhan2013overlooked,mounts2011most} indicate a concerning 26\% error rate in medical imaging analysis during emergency cases.

Computer-assisted diagnosis (CAD) of medical images provides experts (i.e., radiologists, surgeons, etc.) with help in some decision tasks. With the continuous development of deep learning and the improvement of medical image processing techniques \cite{adams2021artificial,choi2020using,chung2018automated,tanzi2020hierarchical}, more and more researchers are trying to employ neural networks for CAD, including fracture detection \cite{bluthgen2020detection,gan2019artificial,kim2018artificial,lindsey2018deep,yahalomi2019detection}.

You Only Look Once (YOLO) model \cite{redmon2016you}, as one of the most important models for object detection tasks \cite{ju2024resolution}, demonstrates satisfactory performance in fracture detection \cite{hrvzic2022fracture,samothai2022evaluation,su2023skeletal}. 
With the introduction of YOLOv8 \cite{jocher2023yolo}, the latest version of the YOLO models, by Ultralytics in 2023, it has been widely used in various object detection tasks. 
As shown in FIGURE \ref{fig_task}, GRAZPEDWRI-DX \cite{nagy2022pediatric} is a public dataset of 20,327 pediatric wrist trauma X-ray images. 
This work follows the study \cite{ju2023fracture} that evaluates the performance of different network models on this dataset in a fracture detection task.

Due to the capacity of attention mechanism to accurately focus on all pertinent information of the input, they are widely applied to various neural network architectures. 
Presently, two primary attention mechanisms exist: spatial attention and channel attention, designed to capture pixel-level pairwise relationships and channel dependencies, respectively \cite{huang2019ccnet,lee2019srm,li2019expectation,zhao2018psanet,zhu2019asymmetric}. 
Studies \cite{cao2019gcnet,fu2019dual,li2019selective,yuan2018ocnet} have demonstrated that incorporating attention mechanism into convolutional blocks shows great potential for performance improvement.

Therefore, we propose the YOLOv8-AM model for fracture detection by employing four different attention modules, including Convolutional Block Attention Module (CBAM) \cite{woo2018cbam}, Global Attention Mechanism (GAM) \cite{liu2021global}, Efficient Channel Attention (ECA) \cite{wang2020eca} and Shuffle Attention (SA) \cite{zhang2021sa}, to the YOLOv8 architecture, as shown in FIGURE \ref{fig_overall}.

Experimental results \cite{woo2018cbam} show that the model performance of combining ResBlock \cite{he2016deep} and CBAM is better than that of CBAM, and therefore we incorporate ResBlock + CBAM (ResCBAM) to the YOLOv8 architecture for experiments. 
After comparing the effects of different attention modules on the YOLOv8-AM model performance, we find that the GAM module has poorer gain effect on the model performance, so we propose ResGAM, which combines ResBlock and GAM, and incorporates this attention module into the YOLOv8 architecture.

The main contributions of this paper are as follows:
\begin{itemize}
\item This work employs four different attention modules to the YOLOv8 architecture and proposes the YOLOv8-AM model for fracture detection, where the YOLOv8-AM model based on ResBlock + CBAM (ResCBAM) achieves the state-of-the-art (SOTA) performance.
\item Since the performance of the YOLOv8-AM model based on GAM is unsatisfactory, we propose ResBlock + GAM (ResGAM) and design the new YOLOv8-AM model based on it for fracture detection.
\item This work demonstrates that compared to the YOLOv8 model, the performances of the YOLOv8-AM models with the incorporation of different attention modules are all greatly improved on the GRAZPEDWRI-DX dataset.  
\end{itemize}

This paper is organized as follows: Section \ref{sec:related} introduces the research on fracture detection utilizing deep learning methods and outlines the evolution of attention mechanism.
Section \ref{sec:method} presents the overall architecture of the YOLOv8-AM model and four different attention modules employed. 
Section \ref{sec:experiment} conducts a comparative analysis of the performance of four YOLOv8-AM models against the baseline YOLOv8 model. 
Subsequently, Section \ref{sec:discussion} discusses the reasons why GAM provides less improvement in the performance of the YOLOv8-AM model for fracture detection. 
Finally, Section \ref{sec:conclusion} concludes this research work and explores the future works.

\section{Related Work}
\label{sec:related}
\subsection{Fracture Detection}
Fracture detection is a hot topic in medical image processing (MIP). 
Researchers usually employ various neural networks for prediction, including the YOLO series models \cite{jocher2023yolo,jocher2020yolov5,redmon2016you}. 
Burkow \emph{et al.} \cite{burkow2022avalanche} utilized the YOLOv5 model \cite{jocher2020yolov5} to recognize rib fracture on 704 pediatric Chest X-ray (CXR) images. 
Tsai \emph{et al.} \cite{tsai2022automatic} performed data augmentation on CXR images and subsequently employed the YOLOv5 model to detect fractures. 
Warin \emph{et al.} \cite{warin2023maxillofacial} firstly categorized the maxillofacial fractures into four categories (i.e., frontal, midfacial, mandibular, and no fracture), and predicted them using the YOLOv5 model on 3,407 CT images. 
In addition, Warin \emph{et al.} \cite{warin2022assessment} used the YOLOv5 model to detect fractures in the X-ray images of the mandible. 
Yuan \emph{et al.} \cite{yuan2021improved} incorporated external attention and 3D feature fusion methods into the YOLOv5 model for fracture detection in skull CT images. Furthermore, vertebral localization proves valuable for recognizing vertebral deformities and fractures. 
Mushtaq \emph{et al.} \cite{mushtaq2022localization} utilized YOLOv5 for lumbar vertebrae localization, achieving the mean Average Precision (mAP) value of 0.957. 
Although the YOLOv5 model is extensively employed in fracture detection, the utilization of the YOLOv8 model \cite{jocher2023yolo} is comparatively rare.

\subsection{Attention Module}
SENet \cite{hu2018squeeze} initially proposed a mechanism to learn channel attention efficiently by applying Global Average Pooling (GAP) to each channel independently. 
Subsequently, channel weights were generated using the Fully Connected layer and the Sigmoid function, leading to the good model performance. 
Following the introduction of feature aggregation and feature recalibration in SENet, some studies \cite{chen20182,gao2019global} attempted to improve the SE block by capturing more sophisticated channel-wise dependencies. 
Woo \emph{et al.} \cite{woo2018cbam} combined the channel attention module with the spatial attention module, introducing the CBAM to improve the representation capabilities of Convolutional Neural Networks (CNNs). 
To reduce information loss and enhance global dimension-interactive features, Liu \emph{et al.} \cite{liu2021global} introduced modifications to CBAM and presented GAM. 
This mechanism reconfigured submodules to magnify salient cross-dimension receptive regions. 
Although these methods \cite{liu2021global,woo2018cbam} have achieved better accuracy, they usually bring higher model complexity and suffer from heavier computational burden. 
Therefore, Wang \emph{et al.} \cite{wang2020eca} proposed the ECA module, which captures local cross-channel interaction by considering every channel and its k neighbors, resulting in significant performance improvement at the cost of fewer parameters.
Different from the ECA module, Zhang \emph{et al.} \cite{zhang2021sa} introduced the SA module. 
This module groups the channel dimensions into multiple sub-features and employs the Shuffle Unit to integrate the complementary sub-features and the spatial attention module for each sub-feature, achieving excellent performance with low model complexity. 
Each of these attention modules can be applied to different neural network architectures to improve model performance.

\begin{figure*}[ht]
  \centering
  \includegraphics[width=\linewidth]{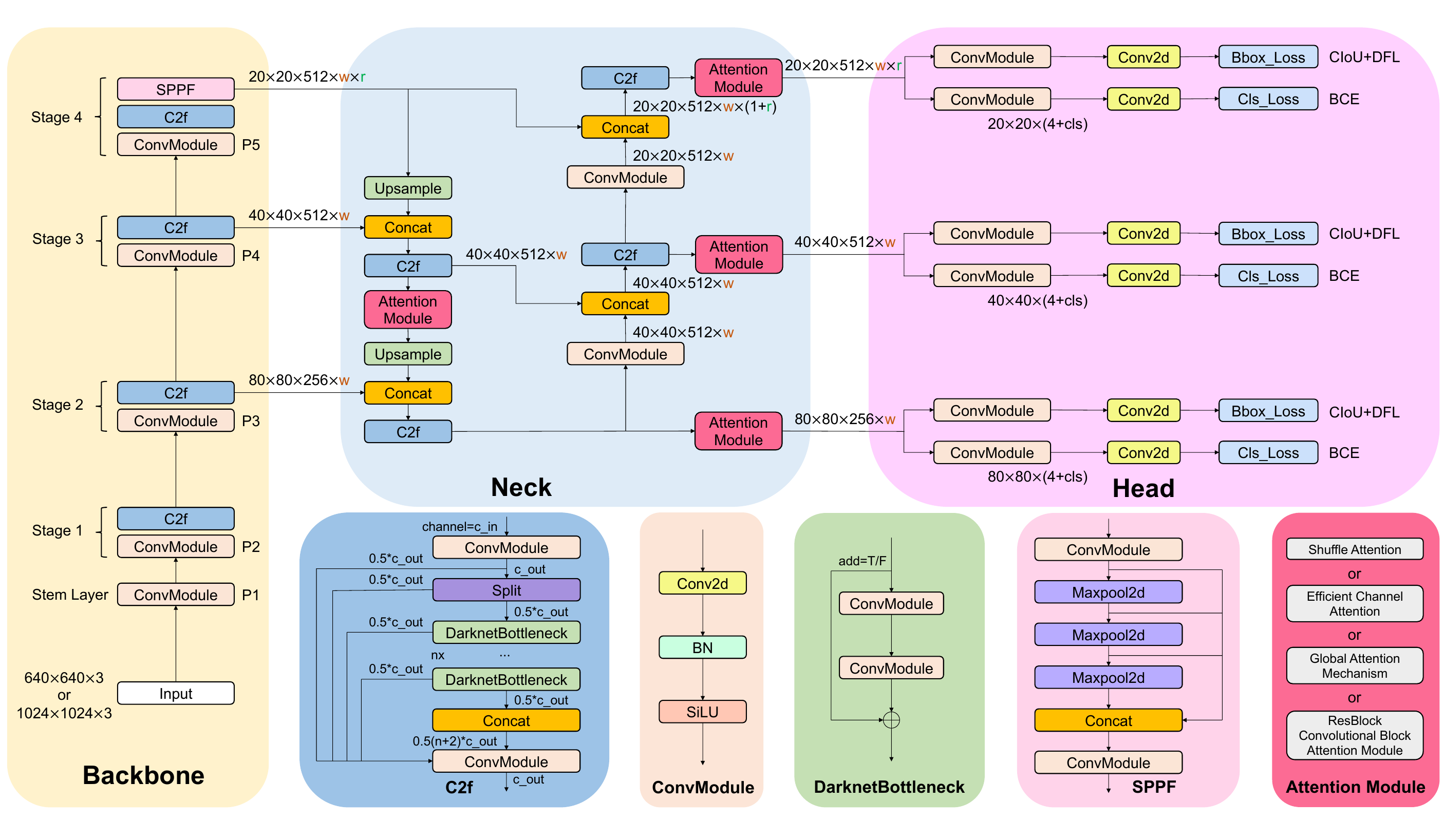}
  \caption{Detailed illustration of the YOLOv8-AM model architecture, where the attention modules are Shuffle Attention (SA), Efficient Channel Attention (ECA), Global Attention Mechanism (GAM), and ResBlock + Convolutional Block Attention Module (ResCBAM), respectively.}
  \label{fig_details}
\end{figure*}

\section{Methodology}
\label{sec:method}
\subsection{Baseline Model}
YOLOv8 architecture comprises four key components: Backbone, Neck, Head, and Loss Function. The Backbone incorporates the Cross Stage Partial (CSP) \cite{wang2020cspnet} concept, offering the advantage of reducing computational loads while enhancing the learning capability of CNNs. Illustrated in FIGURE \ref{fig_details}, YOLOv8 diverges from YOLOv5 employing the C3 module \cite{jocher2020yolov5}, adopting the C2f module, which integrates the C3 module and the Extended ELAN \cite{wang2022designing} (E-ELAN) concept from YOLOv7 \cite{wang2023yolov7}. Specifically, the C3 module involves three convolutional modules and multiple bottlenecks, whereas the C2f module consists of two convolutional modules concatenated with multiple bottlenecks. The convolutional module is structured as Convolution-Batch Normalization-SiLU (CBS).

In the Neck part, YOLOv5 employs the Feature Pyramid Network (FPN) \cite{lin2017feature} architecture for top-down sampling, ensuring that the lower feature map incorporates richer feature information. Simultaneously, the Path Aggregation Network (PAN) \cite{liu2018path} structure is applied for bottom-up sampling, enhancing the top feature map with more precise location information. The combination of these two structures is executed to guarantee the accurate prediction of images across varying dimensions. YOLOv8 follows the FPN and PAN frameworks while deleting the convolution operation during the up-sampling stage, as illustrated in FIGURE \ref{fig_details}.

In contrast to YOLOv5, which employs a coupled head, YOLOv8 adopts a decoupled head to separate the classification and detection heads. Specifically, YOLOv8 eliminates the objectness branch, only retaining the classification and regression branches. Additionally, it departs from anchor-based \cite{ren2015faster} method in favor of anchor-free \cite{duan2019centernet} approach, where the location of the target is determined by its center, and the prediction involves estimating the distance from the center to the boundary.

In YOLOv8, the loss function employed for the classification branch involves the utilization of the Binary Cross-Entropy (BCE) Loss, as expressed by the equation as follows:
\begin{equation}
\label{eq:1}
Loss_{BCE} = -w[y_n \log_{}{x_n}+(1-y_n)\log_{}{(1-x_n)}],
\end{equation}
where $w$ denotes the weight; $y_n$ represents the labeled value, and $x_n$ signifies the predicted value generated by the model.

For the regression branch, YOLOv8 incorporated the use of Distribute Focal Loss (DFL) \cite{li2020generalized} and Complete Intersection over Union (CIoU) Loss \cite{zheng2021enhancing}. The DFL function is designed to emphasize the expansion of probability values around object $y$. Its equation is presented as follows:
\begin{equation}
\begin{split}
\label{eq:2}
Loss_{DF} =&-[(y_{n+1}-y)\log_{}{\frac{y_{n+1}-y_n}{y_{n+1}-y_n}}\\
&+(y-y_n)\log_{}{\frac{y-y_n}{y_{n+1}-y_n}}].
\end{split}
\end{equation}

The CIoU Loss introduces an influence factor to the Distance Intersection over Union (DIoU) Loss \cite{zheng2020distance} by considering the aspect ratio of the predicted bounding box and the Ground-Truth bounding box. The corresponding equation is as follows:
\begin{equation}
\label{eq:3}
Loss_{CIoU} = 1-IoU+\frac{d^2}{c^2}+\frac{v^2}{(1-IoU)+v},
\end{equation}
where $IoU$ measures the overlap between the predicted and Ground-Truth bounding boxes; $d$ is the Euclidean distance between the center points of the predicted and Ground-Truth bounding boxes, and $c$ is the diagonal length of the smallest enclosing box that contains both predicted and Ground-Truth bounding boxes. Additionally, $v$ represents the parameter quantifying the consistency of the aspect ratio, defined by the following equation:
\begin{equation}
\label{eq:4}
v = \frac{4}{\pi^2}(\arctan\frac{w_{gt}}{h_{gt}}-\arctan\frac{w_p}{h_p})^2,
\end{equation}
where $w$ denotes the weight of the bounding box; $h$ represents the height of the bounding box; $gt$ means the Ground-Truth, and $p$ means the prediction.

\begin{figure*}[ht]
  \centering
  \includegraphics[width=\linewidth]{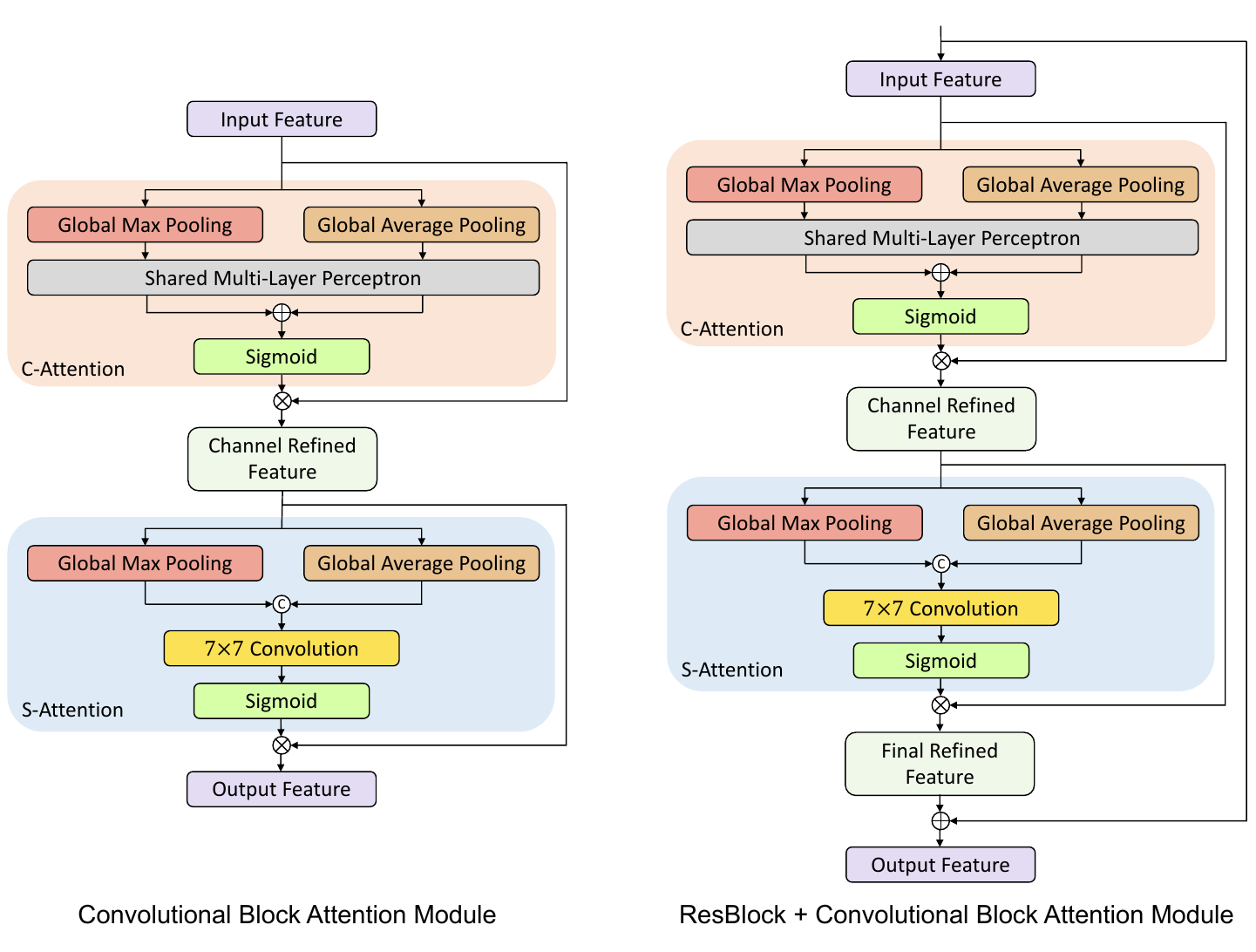}
  \caption{Detailed illustration of Convolutional Block Attention Module (CBAM), the left is the architecture of CBAM, and the right is the architecture of ResCBAM.}
  \label{fig_cbam}
\end{figure*}

\begin{figure*}[ht]
  \centering
  \includegraphics[width=\linewidth]{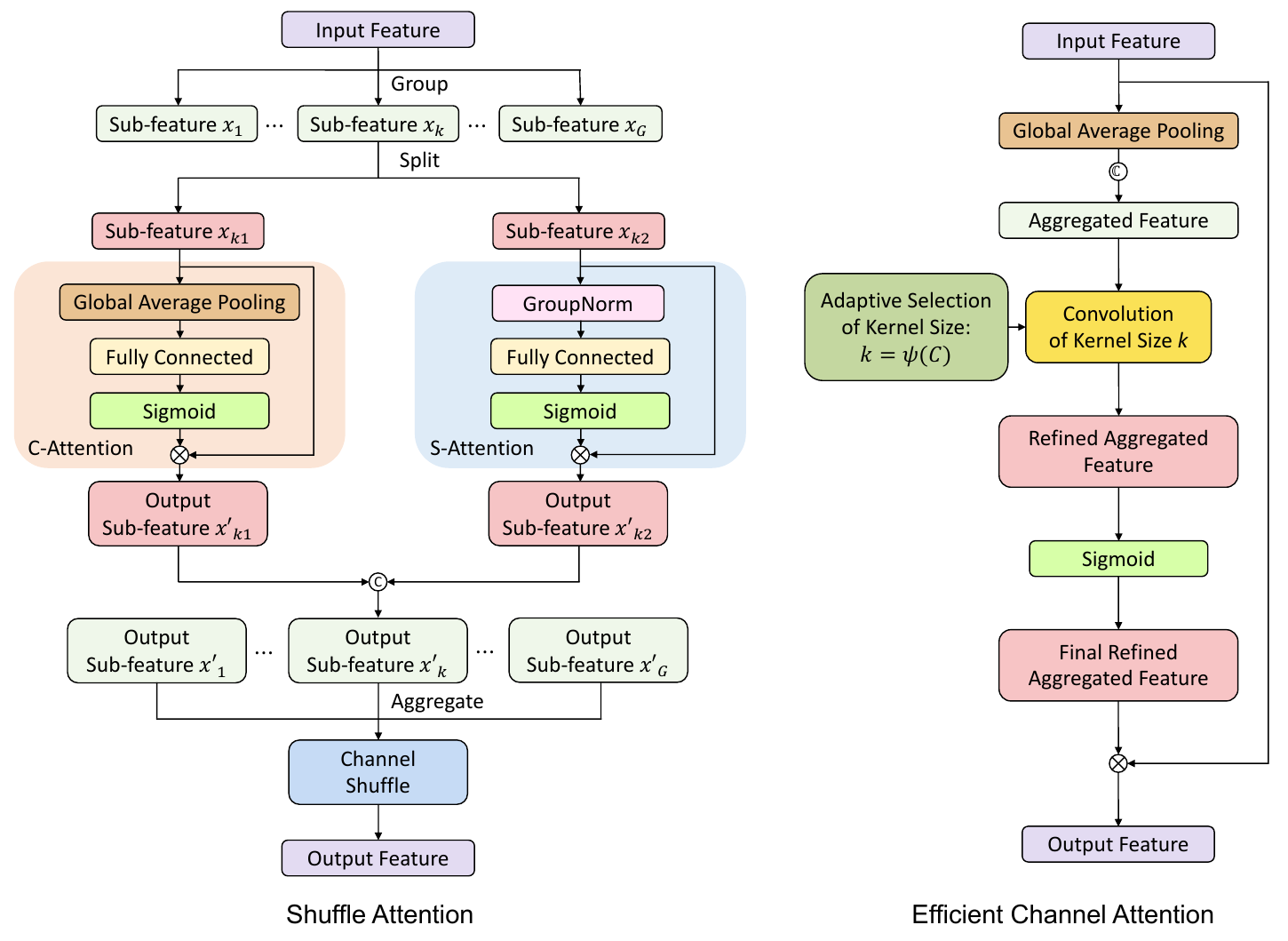}
  \caption{Detailed illustration of Shuffle Attention (SA) and Efficient Channel Attention (ECA), the left is the architecture of SA, and the right is the architecture of ECA.}
  \label{fig_sa_eca}
\end{figure*}

\begin{figure*}[ht]
  \centering
  \includegraphics[width=\linewidth]{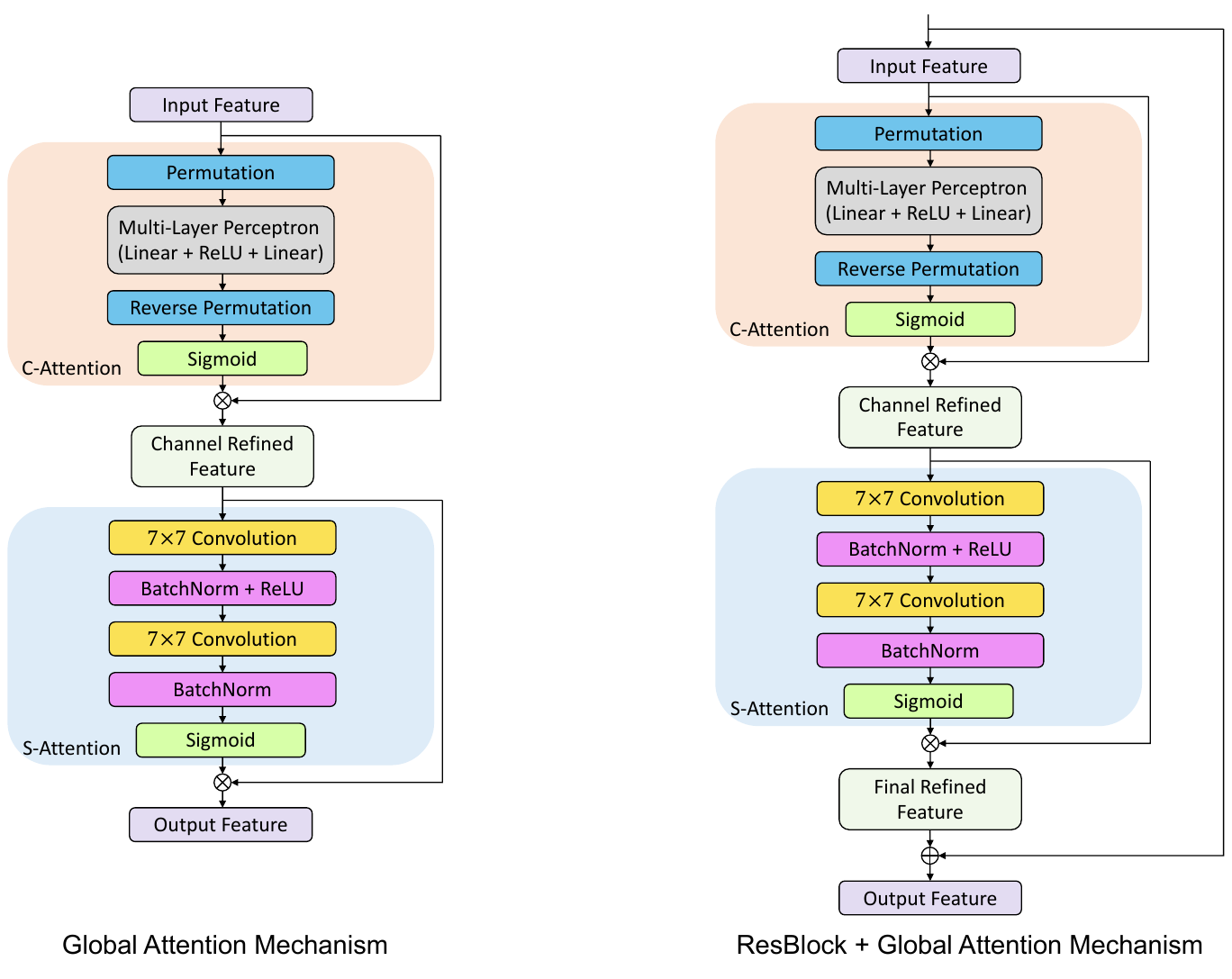}
  \caption{Detailed illustration of Global Attention Mechanism (GAM), the left is the architecture of GAM, and the right is the architecture of ResGAM.}
  \label{fig_gam}
\end{figure*}

\subsection{Proposed Method}
In recent years, the attention mechanism has obtained excellent results in the field of object detection \cite{jiang2022improved,li2020object,zhang2019object}. With the integration of the attention mechanism, the models can recognize the most important information of the input image for extraction and suppress the useless information.

This work incorporates the attention module into the Neck part of YOLOv8 to enhance the capture of key features and suppress the interfering information. As illustrated in FIGURE \ref{fig_details}, attention modules, namely CBAM \cite{woo2018cbam}, GAM \cite{liu2021global}, ECA \cite{wang2020eca}, and SA \cite{zhang2021sa}, are independently employed after each of the four C2f modules. A detailed introduction of these four attention modules is presented in Section \ref{subsec:module}.

\subsection{Attention Modules}
\label{subsec:module}
\subsubsection{Convolutional Block Attention Module (CBAM)}
CBAM comprises both channel attention (C-Attention) and spatial attention (S-Attention), as shown on the left of FIGURE \ref{fig_cbam}. Given an intermediate feature map denoted as $F_{input} \in \mathbb{R}_{C \times H \times W}$, CBAM sequentially infers a 1D channel attention map $M_C \in \mathbb{R}^{C \times 1 \times 1}$ and a 2D spatial attention map $M_S \in \mathbb{R}^{1 \times H \times W}$ through the following equation:
\begin{equation}
\label{eq:5}
F_{CR} = M_C (F_{input}) \otimes F_{input},
\end{equation}
\begin{equation}
\label{eq:6}
F_{FR} = M_S (F_{CR}) \otimes F_{CR},
\end{equation}
where $\otimes$ is the element-wise multiplication; $F_{CR}$ is the Channel Refined Feature, and $F_{FR}$ is the Final Refined Feature. For CBAM, $F_{output}$ is $F_{FR}$ as shown in the following equation:
\begin{equation}
\label{eq:7}
F_{output} = F_{FR}.
\end{equation}

It can be seen from the right of FIGURE \ref{fig_cbam}, for ResBlock + CBAM (ResCBAM), $F_{output}$ is the element-wise summation of $F_{input}$ and $F_{FR}$ as shown in the following equation:
\begin{equation}
\label{eq:8}
F_{output} = F_{input} + F_{FR}.
\end{equation}

Based on the previous studies \cite{zeiler2014visualizing,park2018bam}, CBAM employs both Global Average Pooling (GAP) and Global Max Pooling (GMP) to aggregate the spatial information of a feature map, which generates two different spatial contextual descriptors. Subsequently, these two descriptors share the same Multi-Layer Perceptron (MLP) with one hidden layer. Finally, the output feature vectors from the element-wise summation are input to the sigmoid function ($\sigma$). The specific channel attention equation is as follows:
\begin{equation}
\label{eq:9}
M_C (F)= \sigma[MLP(GAP(F))+MLP(GMP(F))].
\end{equation}

For the spatial attention, CBAM performs GAP and GMP along the channel axis respectively, and then concatenates them together to effectively highlight the information regions \cite{zhou2016learning}, with the symbol $\copyright$ denoting the concatenation. Subsequently, a $7\times7$ convolutional layer is used to perform the convolution operation on these features. The output of this convolution is used as the input of the sigmoid function ($\sigma$). The spatial attention is computed using the following equation:
\begin{equation}
\label{eq:10}
M_S (F)= \sigma[f^{7 \times 7}(GAP(F) \copyright GMP(F))].
\end{equation}

\subsubsection{Efficient Channel Attention (ECA)}
ECA primarily encompasses cross-channel interaction and 1D convolution with an adaptive convolution kernel, as shown on the right of FIGURE \ref{fig_sa_eca}. Cross-channel interaction represents a novel approach to combining features, enhancing the expression of features for specific semantics. The input feature map $F_{input}\in \mathbb{R}^{C \times H \times W}$ obtains the aggregated feature $F_a$ after performing GAP and cross-channel interaction, where $\mathbb{C}$ represents cross-channel interaction. The equation is shown as follows:
\begin{equation}
\label{eq:11}
F_a = \mathbb{C}(GAP(F_{input})).
\end{equation}

For the aggregated features, ECA captures the local cross-channel interaction by considering the interaction between the features of each channel and their neighboring $k$ channels, avoiding the use of 1D convolution for dimensionality reduction, and effectively realizing the multi-channel interaction, where the weights of features $F_{ai}$ are as follows:
\begin{equation}
\label{eq:12}
\omega_i = \sigma(\sum_{j=1}^k \mathbb{W}^j F_{ai}^j), F_{ai}^j \in \Omega_i^k,
\end{equation}
where $\sigma$ is the sigmoid function, and $\Omega_i^k$ denotes the set of $k$ neighboring channels of $F_{ai}$. From Eq. \ref{eq:12} we can see that all the channels have the same inclination parameter, so the model will be more efficient.

For 1D convolution with adaptive convolution kernel, ECA introduces an adaptive method to determine the size of the value. Specifically, the size $k$ of the convolution kernel is directly related to the channel dimension $C$, indicating a nonlinear mapping relationship between them, as illustrated by the following equation:
\begin{equation}
\label{eq:13}
C = \phi(k)=2^{\gamma * k - b}.
\end{equation}

Meanwhile, the convolution kernel size $k$ can be adaptively determined as shown in the following equation:
\begin{equation}
\label{eq:14}
k = \psi(C)=\left|\frac{\log_{2}{C}}{\gamma}+\frac{b}{\gamma}\right|_{odd},
\end{equation}
where $t$ is closest to $\left|t\right|_{odd}$, and based on the experimental results of ECA \cite{wang2020eca}, the values of $\gamma$ and $b$ are set to 2 and 1, respectively.

\subsubsection{Shuffle Attention (SA)}
SA divides the input feature maps into different groups, employing the Shuffle Unit to integrate both channel attention and spatial attention into one block for each group, as shown on the left of FIGURE \ref{fig_sa_eca}. Subsequently, the sub-features are aggregated, and the ``Channel Shuffle`` operator, as employed in ShuffleNetV2 \cite{ma2018shufflenet}, is applied to facilitate information communication among various sub-features.

For channel attention, SA employs GAP to capture and embed global information for the sub-feature $x_{k1}$. In addition, a simple gating mechanism with sigmoid functions is used to create a compact function that facilitates precise and adaptive selection. The final output of the channel attention can be obtained by the following equation:
\begin{equation}
\label{eq:15}
{x_{k1}}' = \sigma [FC(GAP(x_{k1}))]\otimes x_{k1}.
\end{equation}

For spatial attention, to complement the channel attention, SA first uses Group Normalization (GN) \cite{wu2018group} for the sub-feature $x_{k2}$ to obtain spatial-wise statistics. Subsequently, the representation of the output sub-feature ${x_{k2}}'$ is enhanced through the function of $FC()$ as shown in the following equation:
\begin{equation}
\label{eq:16}
{x_{k2}}' = \sigma [FC(GN(x_{k2}))]\otimes x_{k2}.
\end{equation}

It can be seen on the left of FIGURE \ref{fig_sa_eca}, the output sub-feature ${x_k}'$ is obtained by concatenating ${x_{k1}}'$ and ${x_{k2}}'$, and the equation is shown as follows:
\begin{equation}
\label{eq:17}
{x_k}' = {x_{k1}}' \copyright {x_{k2}}'.
\end{equation}

\subsubsection{Global Attention Mechanism (GAM)}
GAM adopts the main architecture proposed by CBAM consisting of channel at-tention and spatial attention, and redesigns the submodules as shown in FIGURE \ref{fig_gam}. Additionally, we add Shortcut Connection \cite{he2016deep} between the layers within GAM, which allows the inputs to propagate forward faster, as shown in the following equation:
\begin{equation}
\begin{split}
\label{eq:18}
F_{output} = F_{input} + &[M_S (M_C (F_{input}) \otimes F_{input})\\
&\otimes (M_C (F_{input}) \otimes F_{input})].
\end{split}
\end{equation}

For channel attention, GAM employs a 3D permutation initially to retain three-dimensional information. Subsequently, it employs a two-layer MLP to amplify the channel-spatial dependencies across dimensions. In summary, the equation is presented as follows:
\begin{equation}
\begin{split}
\label{eq:19}
M_C (F)= \sigma &[Reverse Permutation(\\
&MLP(Permutation(F)))].
\end{split}
\end{equation}

For spatial attention, GAM uses two $7\times7$ convolution layers to integrate spatial information. It also adopts a reduction rate $r$ consistent with the approach in BAM \cite{park2018bam}. The corresponding equation is presented as follows:
\begin{equation}
\label{eq:20}
M_S (F)= \sigma [BN(f^{7\times7} (BN+ReLU(f^{7\times7} (F))))].
\end{equation}

In contrast to CBAM, the authors of GAM considered that max pooling would reduce the amount of information and have a negative effect, so pooling was eliminated to further preserve the feature map.

\begin{table*}[ht]
\centering
\caption{Experimental results of fracture detection on the GRAZPEDWRI-DX dataset using the YOLOv8-AM models with three different attention modules (i.e., ResCBAM, SA, and ECA).}
\begin{subtable}[ht]{\linewidth}
\setlength{\tabcolsep}{17pt}{
\begin{tabular}{ccccccc}
\toprule
\textbf{Model Size} & \textbf{Input Size} & \textbf{mAP$\rm ^{val}_{50}$} & \textbf{mAP$\rm ^{val}_{50-95}$} & \textbf{Params (M)} & \textbf{FLOPs (B)} & \textbf{Inference (ms)} \\ \midrule
Small & 640 & 61.6\% & 38.9\% & 16.06 & 38.27 & 1.9 \\
Medium & 640 & 62.8\% & 39.8\% & 33.84 & 98.19 & 2.9 \\
Large & 640 & 62.9\% & 40.1\% & 53.87 & 196.20 & 4.1 \\ \midrule
Small & 1024 & 63.2\% & 39.0\% & 16.06 & 38.27 & 3.0 \\
Medium & 1024 & 64.3\% & 41.5\% & 33.84 & 98.19 & 5.7 \\
Large & 1024 & 65.8\% & 42.2\% & 53.87 & 196.20 & 8.7 \\ \bottomrule \noalign{\smallskip}
\end{tabular}}
\subcaption{ResBlock + Convolutional Block Attention Module}
\end{subtable}
\begin{subtable}[ht]{\linewidth}
\setlength{\tabcolsep}{17pt}{
\begin{tabular}{ccccccc}
\noalign{\smallskip} \toprule
\textbf{Model Size} & \textbf{Input Size} & \textbf{mAP$\rm ^{val}_{50}$} & \textbf{mAP$\rm ^{val}_{50-95}$} & \textbf{Params (M)} & \textbf{FLOPs (B)} & \textbf{Inference (ms)} \\ \midrule
Small & 640 & 62.7\% & 39.0\% & 11.14 & 28.67 & 1.7 \\
Medium & 640 & 63.3\% & 40.1\% & 25.86 & 79.10 & 2.5 \\
Large & 640 & 64.0\% & 41.5\% & 43.64 & 165.44 & 3.9 \\ \midrule
Small & 1024 & 63.5\% & 39.8\% & 11.14 & 28.67 & 2.9 \\
Medium & 1024 & 64.1\% & 40.3\% & 25.86 & 79.10 & 5.2 \\
Large & 1024 & 64.3\% & 41.6\% & 43.64 & 165.44 & 8.0 \\ \bottomrule \noalign{\smallskip}
\end{tabular}}
\subcaption{Shuffle Attention}
\end{subtable}
\begin{subtable}[ht]{\linewidth}
\setlength{\tabcolsep}{17pt}{
\begin{tabular}{ccccccc}
\noalign{\smallskip} \toprule
\textbf{Model Size} & \textbf{Input Size} & \textbf{mAP$\rm ^{val}_{50}$} & \textbf{mAP$\rm ^{val}_{50-95}$} & \textbf{Params (M)} & \textbf{FLOPs (B)} & \textbf{Inference (ms)} \\ \midrule
Small & 640 & 61.4\% & 37.4\% & 11.14 & 28.67 & 1.9 \\
Medium & 640 & 62.1\% & 38.7\% & 25.86 & 79.10 & 2.5 \\
Large & 640 & 62.6\% & 40.2\% & 43.64 & 165.45 & 3.6 \\ \midrule
Small & 1024 & 62.1\% & 38.7\% & 11.14 & 28.67 & 2.7 \\
Medium & 1024 & 62.4\% & 40.1\% & 25.86 & 79.10 & 5.2 \\
Large & 1024 & 64.2\% & 41.9\% & 43.64 & 165.45 & 7.7 \\ \bottomrule \noalign{\smallskip}
\end{tabular}}
\subcaption{Efficient Channel Attention}
\end{subtable}
\label{tab:1}
\end{table*}

\begin{table*}[ht]
\centering
\caption{Experimental results of fracture detection on the GRAZPEDWRI-DX dataset using the YOLOv8-AM models with two attention modules (i.e., GAM and ResGAM).}
\begin{subtable}[ht]{\linewidth}
\setlength{\tabcolsep}{17pt}{
\begin{tabular}{ccccccc}
\toprule
\textbf{Model Size} & \textbf{Input Size} & \textbf{mAP$\rm ^{val}_{50}$} & \textbf{mAP$\rm ^{val}_{50-95}$} & \textbf{Params (M)} & \textbf{FLOPs (B)} & \textbf{Inference (ms)} \\ \midrule
Small & 640 & 0.625 & 0.397 & 13.86 & 34.24 & 2.2 \\
Medium & 640 & 0.628 & 0.398 & 30.27 & 90.26 & 3.6 \\
Large & 640 & 0.633 & 0.407 & 49.29 & 183.53 & 8.7 \\ \midrule
Small & 1024 & 0.635 & 0.400 & 13.86 & 34.24 & 4.3 \\
Medium & 1024 & 0.637 & 0.405 & 30.27 & 90.26 & 8.9 \\
Large & 1024 & 0.642 & 0.410 & 49.29 & 183.53 & 12.7 \\ \bottomrule \noalign{\smallskip}
\end{tabular}}
\subcaption{Global Attention Mechanism}
\end{subtable}
\begin{subtable}[ht]{\linewidth}
\setlength{\tabcolsep}{17pt}{
\begin{tabular}{ccccccc}
\noalign{\smallskip} \toprule
\textbf{Model Size} & \textbf{Input Size} & \textbf{mAP$\rm ^{val}_{50}$} & \textbf{mAP$\rm ^{val}_{50-95}$} & \textbf{Params (M)} & \textbf{FLOPs (B)} & \textbf{Inference (ms)} \\ \midrule
Small & 640 & 61.4\% & 38.6\% & 13.86 & 34.24 & 2.7 \\
Medium & 640 & 62.8\% & 40.5\% & 30.27 & 90.26 & 3.9 \\
Large & 640 & 64.0\% & 41.2\% & 49.29 & 183.53 & 9.4 \\ \midrule
Small & 1024 & 64.8\% & 41.2\% & 13.86 & 34.24 & 4.4 \\
Medium & 1024 & 64.9\% & 41.3\% & 30.27 & 90.26 & 12.4 \\
Large & 1024 & 65.0\% & 41.8\% & 49.29 & 183.53 & 18.1 \\ \bottomrule \noalign{\smallskip}
\end{tabular}}
\subcaption{ResBlock + Global Attention Mechanism}
\end{subtable}
\label{tab:2}
\end{table*}

\begin{table*}[ht]
\centering
\caption{Quantitative comparison (F1-Score/mAP/Inference) of fracture detection on the GRAZPEDWRI-DX dataset using YOLOv8 and YOLOv8-AM models. Best and 2nd best performance are in {\color{red}red} and {\color{blue}blue} colors, respectively.}
\setlength{\tabcolsep}{21pt}{
\begin{tabular}{ccccccc}
\toprule
\textbf{Module} & \textbf{N/A} & \textbf{ResCBAM} & \textbf{SA} & \textbf{ECA} & \textbf{GAM} & \textbf{ResGAM} \\ \midrule
Params & 43.61M & 53.87M & 43.64M & 43.64M & 49.29M & 49.29M \\
FLOPs & 164.9B & 196.2B & 165.4B & 165.5B & 183.5B & 183.5B \\
F1-Score & 0.62 & {\color{blue}0.64} & 0.63 & {\color{red}0.65} & {\color{red}0.65} & {\color{blue}0.64} \\ 
mAP$\rm ^{val}_{50}$ & 63.6\% & {\color{red}65.8\%} & 64.3\% & 64.2\% & 64.2\% & {\color{blue}65.0\%} \\
mAP$\rm ^{val}_{50-95}$ & 40.4\% & {\color{red}42.2\%} & 41.6\% & {\color{blue}41.9\%} & 41.0\% & 41.8\% \\ 
Inference & {\color{red}7.7ms} & 8.7ms & {\color{blue}8.0ms} & {\color{red}7.7ms} & 12.7ms & 18.1ms \\ \bottomrule
\end{tabular}}
\label{tab:3}
\end{table*}

\begin{figure*}[ht]
  \centering
  \includegraphics[width=\linewidth]{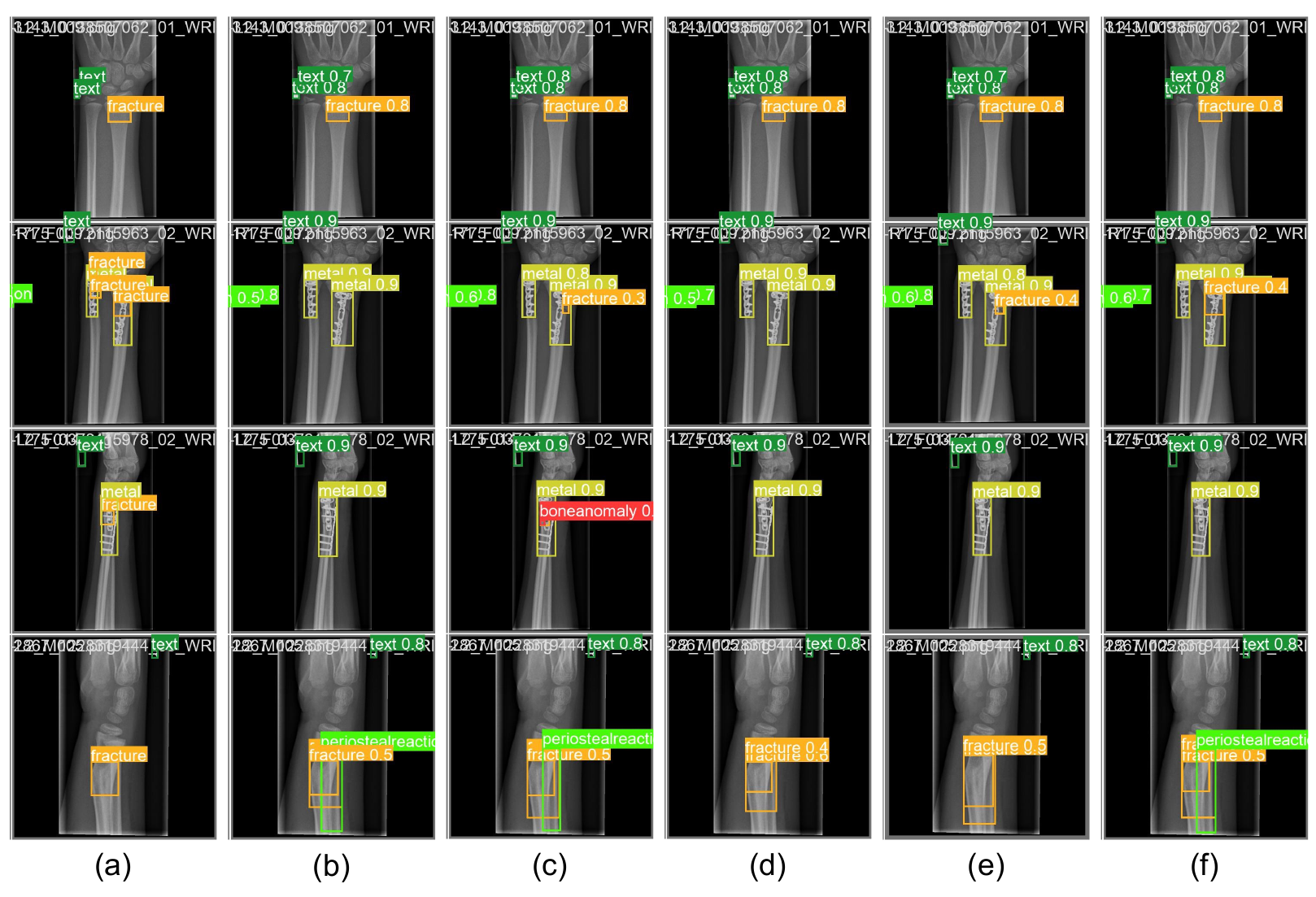}
  \caption{Examples of results of different YOLOv8-AM models applied to pediatric wrist fracture detection and Ground-Truth, where (a) manually labeled data; (b) ResCBAM; (c) ECA; (d) SA; (e) GAM; and (f) ResGAM.}
  \label{fig_prediction}
\end{figure*}

\begin{figure*}[ht]
  \centering
  \includegraphics[width=\linewidth]{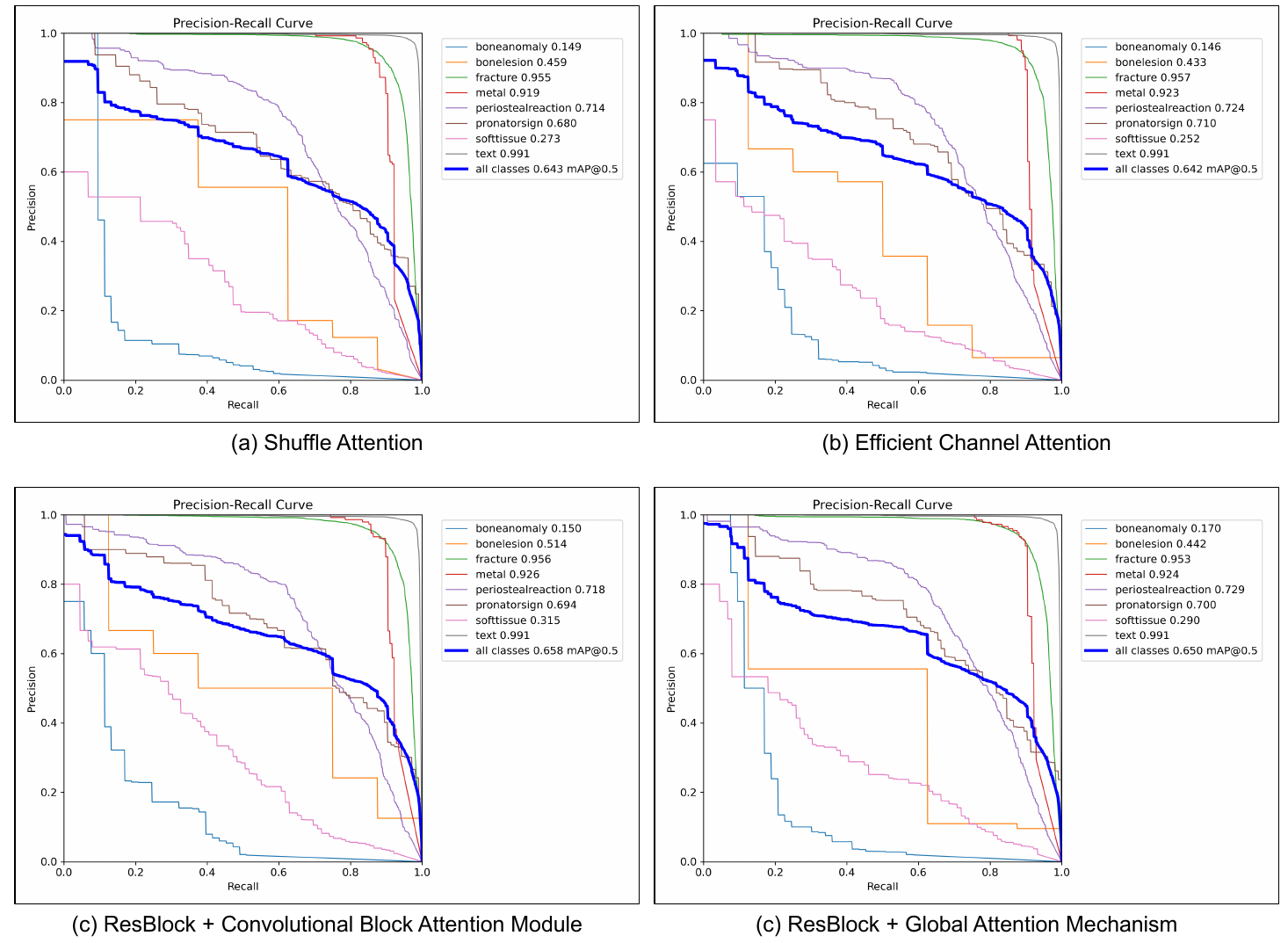}
  \caption{Detailed illustrations of the Precision-Recall Curves (\%) for four different YOLOv8-AM models on each category in the GRAZPEDWRI-DX dataset, where (a) SA; (b) ECA; (c) ResCBAM; and (d) ResGAM.}
  \label{fig_pr}
\end{figure*}

\section{Experiment}
\label{sec:experiment}
\subsection{Dataset}
GRAZPEDWRI-DX \cite{nagy2022pediatric} is a public dataset of 20,327 pediatric wrist trauma X-ray images released by the University of Medicine of Graz. These X-ray images were collected by multiple pediatric radiologists at the Department for Pediatric Surgery of the University Hospital Graz between 2008 and 2018, involving 6,091 patients and a total of 10,643 studies. This dataset is annotated with 74,459 image labels, featuring a total of 67,771 labeled objects. 

\subsection{Preprocessing and Data Augmentation}
In the absence of predefined training, validation, and test sets by the dataset publisher, we perform a random division, allocating 70\% to the training set, 20\% to the validation set, and 10\% to the test set. Specifically, the training set comprises 14,204 images (69.88\%), the validation set includes 4,094 images (20.14\%), and the test set comprises 2,029 images (9.98\%).

Due to the limited diversity in brightness among the X-ray images within the GRAZPEDWRI-DX dataset, the model trained only on these images may not perform well in predicting other X-ray images. Therefore, to enhance the robustness of the model, we employ data augmentation techniques to expand the training set. More specifically, we fine-tune the contrast and brightness of the X-ray images using the addWeighted function available in OpenCV, which is an open-source computer vision library.

\subsection{Evaluation Metric}
\subsubsection{Parameters (Params)}
The quantity of parameters in a model depends on its architectural complexity, the number of layers, neurons per layer, and various other factors. More parameters in a model usually means a larger model size. Generally, the larger the model obtains the better model performance, but it also means that more data and computational resources are needed for training. In real-world applications, it is necessary to balance the relationship between model complexity and computational cost.

\subsubsection{Floating Point Operations (FLOPs)}
Floating Point Operations serves as a metric for assessing the performance of computers or computing systems and is commonly employed to evaluate the computational complexity of neural network models. FLOPs represent the number of floating-point operations executed per second, providing a crucial indicator of the computational performance and speed of the model. In resource-limited environments, models with lower FLOPs may be more suitable, while those with higher FLOPs may necessitate more powerful hardware support.

\subsubsection{Mean Average Precision (mAP)}
Mean Average Precision is a common metric used to evaluate the performance of object detection models. In the object detection task, the prediction aim of the model is to recognize objects within the image and determine their locations. Precision measures the proportion of objects detected by the model that correspond to the real objects, while recall measures the proportion of the real objects detected by the model. These two metrics need to be weighed, and mAP is a combination of precision and recall.

For each category, the model computes the area under the Precision-Recall curve, referred to as Average Precision. This metric reflects the performances of models for each category. We subsequently average the computed Average Precision values across all categories to derive the overall mAP.

\subsubsection{F1-Score}
The F1-Score is based on the summed average of the precision and recall values of the model. Its value ranges from 0 to 1, where value closer to 1 indicating that the model has a better balance between precision and recall. When one of precision and recall values is biased towards 0, the F1-Score will also be close to 0, which indicates poor model performance. Considering both precision and recall values together, the F1-Score helps to assess the accuracy of the model in positive category prediction and its sensitivity to positive categories.

In the problem involving imbalanced category classification, relying solely on accuracy for evaluation may lead to bias, as the model may perform better in predicting the category with a larger number of samples. Therefore, using the F1-Score provides a more comprehensive assessment of the model performance, especially when dealing with imbalanced datasets.

\subsubsection{Inference Time}
Inference Time refers to the duration required by network models to process X-ray images from input to the final prediction, including preprocessing, inference, and post-processing stages. 
In this work, the inference time per image is measured using single NVIDIA GeForce RTX 3090 GPU.

\subsection{Experiment Setup}
We train the YOLOv8 model and different YOLOv8-AM models on the dataset \cite{nagy2022pediatric}. In contrast to the 300 epochs recommended by Ultralytics \cite{jocher2023yolo} for YOLOv8 training, the experimental results \cite{ju2023fracture} indicate that the best performance is achieved within 60 to 70 epochs. Consequently, we set 100 epochs for all models training.

For the hyperparameters of model training, we select the SGD \cite{ruder2016overview} optimizer instead of the Adam \cite{kingma2014adam} optimizer based on the result of the ablation experiment in study \cite{ju2023fracture}. Following the recommendation of Ultralytics \cite{jocher2023yolo}, this work establishes the weight decay of the optimizer at 5e-4, coupled with a momentum of 0.937, and the initial learning rate to 1e-2. To compare the effects of different input image sizes on the performance of the models, this work sets the input image size to 640 and 1024 for the experiments respectively.
This work employs Python 3.9 for training all models on the framework of PyTorch 1.13.1. We advise readers to utilize versions higher than Python 3.7 and PyTorch 1.7 for model training, and the specific required environment can be accessed on our GitHub repository. All experiments are executed using one single NVIDIA GeForce RTX 3090 GPU, with the batch size of 16 set to accommodate GPU memory constraints.

\subsection{Experimental Results}
In the fracture detection task, to compare the effect of different input image sizes on the performance of the YOLOv8-AM model, we train our model using training sets with input image sizes of 640 and 1024, respectively. Subsequently, we evaluate the performance of the YOLOv8-AM model based on different attention modules on the test set with the corresponding image sizes.

As shown in TABLEs \ref{tab:1} and \ref{tab:2}, the performance of the models trained using the training set with input image size of 1024 surpasses that of models trained using the training set with input image size of 640. Nevertheless, it is noteworthy that this improvement in performance is accompanied by an increase in inference time. For example, considering the ResCBAM-based YOLOv8-AM model with the large model size, the mean Average Precision at IoU 50 (mAP 50) attains 42.2\% for the input image size of 1024, which is 5.24\% higher than that of 40.1\% obtained for the input image size of 640. However, the inference time of the model increases from 4.1ms to 8.7ms because the model size becomes larger.

TABLE \ref{tab:1} shows the model performance of the YOLOv8-AM, employing three different attention modules including ResCBAM, SA and ECA. These experimental results are obtained with different model sizes and different input image sizes. In TABLE \ref{tab:2}, the model performance of the YOLOv8-AM is presented when utilizing GAM directly. Additionally, we introduce a novel approach by incorporating ResBlock and GAM, named ResGAM, to enhance the overall model performance of the YOLOv8-AM. Specifically, when the input image size is 1024 and the model size is medium, the newly introduced ResGAM demonstrates a notable enhancement in mAP for the YOLOv8-AM model based on GAM. The mAP 50 increases from 63.7\% to 64.9\%, providing that our proposed ResGAM positively contributes to the performance enhancement.

To compare the effect of different attention modules on the model performance, we organize the experimental data with input image size of 1024 and model size of large. The corresponding results are presented in TABLE \ref{tab:3}. In summary, the F1-Score, mAP 50-95, and mAP 50 values for all YOLOv8-AM models surpass that of the YOLOv8 model. Specifically, the mAP 50 for the YOLOv8-AM models based on SA and ECA stands at 64.3\% and 64.2\%, respectively. These values are marginally superior to 63.6\% mAP 50 obtained by the YOLOv8 model. Notably, this enhanced performance requires almost the same inference time as that of the YOLOv8 model. For the YOLOv8-AM model based on ResCBAM, it obtained mAP 50 of 65.8\%, achieving the SOTA model performance. However, the gain on the performance of the YOLOv8 model by GAM is not satisfactory, so we propose the incorporation of ResGAM into the YOLOv8-AM model.

In this paper, to evaluate the gain of the attention module on the accuracy of the YOLOv8 model for predicting fractures in a real-world diagnostic scenario, four X-ray images are randomly selected, and FIGURE \ref{fig_prediction} demonstrates the prediction results of different YOLOv8-AM models. The YOLOv8-AM model, serving as a CAD tool, plays a crucial role in supporting radiologists and surgeons during diagnosis by effectively identifying fractures and detecting metal punctures in singular fracture scenarios. However, it is important to note that the accuracy of model prediction may decrease in instances involving dense fractures.

FIGURE \ref{fig_pr} shows the Precision-Recall Curve (PRC) for each category predicted by different YOLOv8-AM models. From the figure, we can see that different YOLOv8-AM models have greater ability to correctly detect fracture, metal, and text, with the average accuracy exceeding 90\%. However, for two categories, bone anomaly and soft tissue, the ability to correctly detect them is poorer, with accuracy approximately at 45\% and 30\%, respectively. These low accuracy rates seriously effect the mAP 50 values of the models. We consider this is due to the small number of objects within these two categories in the used dataset. As described in GRAZPEDWRI-DX \cite{nagy2022pediatric}, the number of bone anomaly and soft tissue accounts for 0.41\% and 0.68\% of the total number of objects, respectively. Consequently, any improvement in model performance via architectural enhancements is constrained by this data limitation. To enhance the performance of the model, a recourse to incorporating extra data becomes imperative.

\section{Discussion}
\label{sec:discussion}
It is evident from TABLE \ref{tab:3} that the gain of GAM on the model performance of the YOLOv8-AM is poor on the GRAZPEDWRI-DX dataset. To further enhance the performance of the YOLOv8-AM model based on GAM, we have designed ResGAM, but it is still not as good as the performance gain provided by ResCBAM. According to the related theory \cite{vaswani2017attention}, we think this is due to the decision of GAM to abandon pooling. In the attention mechanism, pooling serves to extract crucial features in each channel, thereby facilitating a concentration on the important aspects.

\cite{liu2021global} demonstrated performance enhancements surpassing those achieved with CBAM on the CIFAR100 \cite{krizhevsky2009learning} and ImageNet-1K \cite{deng2009imagenet} datasets by enabling the neural network to acquire features across all dimensions, leveraging the remarkable adaptability inherent in the neural network. Nevertheless, it is noteworthy that the CIFAR100 dataset comprises 50,000 images representing diverse scenes in its training set, while the training set of the ImageNet-1K dataset includes a total of 1,281,167 images. In contrast, our model is trained using a small training set of 14,204 X-ray images. Consequently, the neural network is only required to learn the important features, such as bone fractures and lesions, within the X-ray images. This situation is different from the theory proposed by \cite{liu2021global}, given the limited scope of our dataset and the specific focus on relevant features on X-ray images.

\section{Conclusion and Future Works}
\label{sec:conclusion}
Following the introduction of the YOLOv8 model by Ultralytics in 2023, researchers began to employ it for the detection of fractures across various parts of the body. While the YOLOv8 model, the latest version of the YOLO models, demonstrated commendable performance on the GRAZPEDWRI-DX dataset, it fell short of achieving the SOTA. To address this limitation, we incorporate four attention modules (CBAM, ECA, SA, and GAM) into the YOLOv8 architecture respectively to enhance the model performances. Additionally, we combine ResBlock with CBAM and GAM to form ResCBAM and ResGAM, respectively. Notably, the mAP 50 for the YOLOv8-AM model based on ResGAM improves from 64.2\% (GAM) to 65.0\% without increasing the model Parameters and FLOPs. Meanwhile, the mAP 50 for the YOLOv8-AM model with ResCBAM obtains a superior performance of 65.8\%, surpassing the SOTA benchmark.

To support this work's application as CAD tools in medical imaging diagnosis, we plan to deploy the proposed YOLOv8-AM models as web and mobile applications (e.g., Android and iOS). This will ensure ease of use for both surgeons and medical professionals.

\bibliographystyle{IEEEtran}
\bibliography{bibliography}
\EOD
\end{document}